\pdfoutput=1

\documentclass[acmsmall]{acmart}
\usepackage{multirow}
\usepackage{xcolor}
\newcommand{\lhr}[1]{\textcolor{black}{#1}}

\usepackage{amsmath}
\usepackage{tabularx}

\AtBeginDocument{%
  }

\setcopyright{acmlicensed}
\copyrightyear{2024}
\acmYear{2024}
\acmDOI{XXXXXXX.XXXXXXX}

\acmJournal{JACM}
\acmVolume{37}
\acmNumber{4}
\acmArticle{111}
\acmMonth{8}

\begin{document}

\title{Privacy in Large Language Models: Attacks, Defenses and Future Directions}

\author{Haoran Li}
\authornote{Haoran Li, Yulin Chen and Jinglong Luo contribute equally.}
\email{hlibt@conncet.ust.hk}
\affiliation{%
  \institution{HKUST}
  \country{Hong Kong SAR, China}
}

\author{Yulin Chen}
\authornotemark[1]
\affiliation{%
  \institution{National University of Singapore}
  \country{Singapore}}
\email{chenyulin28@u.nus.edu}

\author{Jinglong Luo}
\authornotemark[1]
\affiliation{%
  \institution{Harbin Institute of Technology, Shenzhen and Peng Cheng Lab}
  \city{Shenzhen}
  \country{China}}
\email{jinglongluo1@gmail.com}

\author{Jiecong Wang}
\affiliation{%
  \institution{Beihang University}
  \city{Beijing}
  \country{China}}
\email{jcwang@buaa.edu.cn}

\author{Hao Peng}
\affiliation{%
  \institution{Beihang University}
  \city{Beijing}
  \country{China}}
\email{penghao@buaa.edu.cn}

\author{Yan Kang}
\affiliation{%
  \institution{Webank}
  \city{Shenzhen}
  \country{China}}
\email{kangyan2003@gmail.com}

\author{Xiaojin Zhang}
\affiliation{%
  \institution{Huazhong University of Science and Technology}
  \country{China}}
\email{xiaojinzhang@hust.edu.cn}

\author{Qi Hu}
\email{qhuaf@connect.ust.hk}
\affiliation{%
  \institution{HKUST}
  \country{Hong Kong SAR, China}
}

\author{Chunkit Chan}
\email{ckchancc@connect.ust.hk}
\affiliation{%
  \institution{HKUST}
  \country{Hong Kong SAR, China}
}

\author{Zenglin Xu}
\email{zenglinxu@fudan.edu.cn}
\affiliation{%
  \institution{Fudan University and Peng Cheng Lab}
  \city{Shanghai}
  \country{China}}

\author{Bryan Hooi}
\email{bhooi@comp.nus.edu.sg}
\affiliation{%
  \institution{National University of Singapore}
  \country{Singapore}
}

\author{Yangqiu Song}
\email{yqsong@cse.ust.hk}
\affiliation{%
  \institution{HKUST}
  \country{Hong Kong SAR, China}
}

\renewcommand{\shortauthors}{Li et al.}

\begin{abstract}
With the advancement of deep learning and transformer models, large language models (LLMs) have significantly enhanced the ability to effectively tackle various downstream NLP tasks and unify these tasks into generative pipelines. 
On the one hand,  powerful language models, trained on massive textual data, have brought unparalleled accessibility and usability for both models and users.
These LLMs have significantly lowered the entry barrier for application developers and users, as they provide pre-trained language understanding and instruction-following capabilities.
The availability of powerful LLMs has opened up new possibilities across various fields, including LLM-enabled agents, virtual assistants, chatbots, and more.
On the other hand, unrestricted access to these models can also introduce potential malicious and unintentional privacy risks. 
The same capabilities that make these models valuable tools can also be exploited for malicious purposes or unintentionally compromise sensitive information. 
Despite ongoing efforts to address the safety and privacy concerns associated with LLMs, the problem remains unresolved.
In this paper, we aim to offer a thorough examination of the current privacy attacks targeting LLMs and categorize them according to the adversary's assumed capabilities to shed light on the potential vulnerabilities presented in LLMs.
Then, we delve into an exploration of prominent defense strategies that have been developed to mitigate the risks of these privacy attacks. 
In addition to discussing existing works, we also address the upcoming privacy concerns that may arise as these LLMs continue to evolve.
Lastly, we conclude our paper by highlighting several promising directions for future research and exploration in the field of LLM privacy.
By identifying these research directions, we aim to inspire further advancements in privacy protection for LLMs and contribute to more secure and privacy-aware development of these powerful LLMs.
With this survey, we hope to provide valuable insights into the potential vulnerabilities that exist within LLMs, thus highlighting the importance of addressing privacy concerns in their development and applications.
\end{abstract}




\maketitle

\section{Introduction}

\subsection{Motivation}
With the development of deep transformer models in natural language processing, pre-trained language models (LMs) mark the beginning of a transformative era for natural language processing and society as a whole.
Presently,  generative large language models (LLMs) present remarkable capabilities by combining various natural language processing tasks into a comprehensive text generation framework.
These models, such as OpenAI's GPT-4, Anthropic's Claude 2 and Meta's Llama 2, have made significant impacts in recent years for understanding and generating human language.
As a result, these LLMs achieve unparalleled performance on both predefined tasks and real-world challenges~\cite{2020t5,2022flant5,Brown2020LanguageMA,OpenAI2023GPT4TR, ouyang2022training, DBLP:journals/corr/abs-2305-12870, chan2023chatgpt}.
Besides generating coherent and contextually relevant text across various applications, LLMs can automate many language-related tasks, making them invaluable tools for developers and end-users.
Furthermore, LLMs have the ability to generalize to vast unseen corpora of text. With proper instructions (prompts) and demonstrations, LLMs can even adapt to specific contexts or tackle novel tasks without further tuning~\cite{Chen2021EvaluatingLL, zhou2023leasttomost, Kojima2022LargeLM, Wei2022ChainOT,sanh2022multitask}.
Thus, it has become trending to integrate LLMs into various applications, from scientific research to smart assistants.

In addition to the enhanced performance, the training data scale of language models also expands along with the models' sizes.
These LLMs are not solely trained on annotated textual data for specific tasks, but they also devour a vast amount of public textual data from the Internet.
Unlike meticulously curated annotation data, the free-form texts extracted from the Internet suffer from poor data quality and inadvertent personal information leakage.
For instance, simple interactions with the models may incur accidental dissemination of personally identifiable information (PII) ~\cite{li2023multi, Lukas2023AnalyzingLO, huang-etal-2022-large,carlini-2021-extracting,zou2023universal,wang2023decodingtrust}.
Unfortunately, unintended PII exposure without the awareness or consent of the individuals involved may result in violations of existing privacy laws, such as the EU's General Data Protection Regulation (GDPR) and the California Consumer Privacy Act (CCPA).
Moreover, integrating diverse applications into LLMs is a growing trend aimed at enhancing their knowledge grounding capabilities. 
These integrations enable LLMs to effectively solve math problems (such as ChatGPT + Wolfram Alpha), read formatted files (like ChatPDF), and provide responses to users' queries using search engines (such as the New Bing).
When LLMs are combined with external tools like search engines, additional domain-specific privacy and security vulnerabilities emerge. 
For instance, as reported in \citet{li2023multi}, a malicious adversary may exploit the New Bing to associate the victims' PII even given their partial information.
Consequently, the extent of privacy breaches in the present-day LLMs remains uncertain.

To ensure the privacy of data subjects, various privacy protection mechanisms have been proposed. 
In particular, several studies~\cite{qu2021natural,yue2022synthetic,yu2022differentially,Igamberdiev-2023-DP-BART,Li2023PBenchAM} exploited differential privacy (DP)~\cite{Dwork-08-DP} to safeguard data subjects' privacy during LLMs' sensitive data fine-tuning stages.
While DP offers a theoretical worst-case privacy guarantee for the protected data, current privacy mechanisms significantly compromise the utility of LLMs, rendering many existing approaches impractical.
Thus, to develop LLMs for public benefit, it is empirical to study the trade-off between privacy and utility.
Cryptography-based LLMs~\cite{chen2022x, li2022mpcformer, liang2023merge,hao2022iron, zheng2023primer, dong2023puma, hou2023ciphergpt, ding2023east} refers to methods with cryptography techniques, such as Secure Multi-Party Computation (SMPC) and Homomorphic Encryption (HE).
Despite its impressive growth and effectiveness in enhancing privacy during training, Cryptography-based LLMs still encounters challenges in areas such as privacy-preserving inference and the adaptability of models.
Federated learning (FL), a privacy-focused distributed learning approach, allows multiple entities to collaboratively train or refine their language models without exchanging the private data held by each data owner. Although FL is designed to safeguard data privacy by obstructing direct access to private data by potential adversaries, research indicates that FL algorithms are still vulnerable to data privacy breaches even with privacy safeguards. Such breaches can occur through data inference attacks conducted by either \textit{semi-honest}~\cite{Zhu-2019-Deep,Zhao-2020-iDLG,yin2021see,geiping2020inverting,gupta2022recovering,balunovic2022lamp} or \textit{malicious adversaries}~\cite{fowl2023decepticons,chu2023panning}.
To address the aforementioned challenges, it is crucial to gain a clear understanding of privacy in the context of LLMs. 
Instead of discussing the broad concept of privacy, in this paper, we will delve into the concept of privacy by comprehensively exploring and analyzing the existing privacy attacks and defenses that are applicable to LLMs. 
After the analysis, we point out the future directions to achieve the privacy-preserving LLMs.

\subsection{Scope and objectives}
This survey examines recent advancements in privacy attacks and defenses on LLMs.
In comparison to several recent survey papers~\cite{Brown-2022-what,Ishihara2023TrainingDE,Mozes2023UseOL,cheng2023backdoor} about privacy in LLMs, our work offers a more comprehensive and systematic analysis. 
We go beyond previous surveys by incorporating the most recent advancements in LLMs, ensuring that our analysis is up-to-date and relevant.
Furthermore, we also investigate novel techniques and strategies that have emerged to safeguard user privacy, such as differential privacy~\cite{Dwork-08-DP}, Cryptography-based methods~\cite{Mohassel17_MPC}, unlearning, and federated learning~\cite{YangLCT19,konevcny2016federated,lifedassistant}.
By evaluating these defense mechanisms, we aim to provide valuable insights into their effectiveness and limitations.
Finally, after analyzing the attacks and defenses, we discuss future unstudied privacy vulnerabilities and potential remedies to solve the problem.

\subsection{Organization}
The whole paper is organized in the following sections.
Section~\ref{ch:background} introduces preliminary knowledge about the fundamental concepts of privacy and language models.
Section~\ref{ch:Privacy Attacks} presents a concise overview of the various privacy attacks targeting LLMs.
Section~\ref{ch:Privacy Defenses} examines the current defense mechanisms designed to safeguard the data privacy of LLMs, along with their inherent limitations.
Section~\ref{ch:Future} enumerates several potential privacy vulnerabilities and proposes future research directions for developing defense mechanisms.
Lastly, Section~\ref{ch:Conclusion} concludes the aforementioned content.
\section{Backgrounds}\label{ch:background}
In this section, we introduce preliminary knowledge about LLMs and privacy.
Firstly, we briefly discuss LLMs' development over the recent years.
Secondly, we talk about differential privacy, a probabilistic view of information privacy during data aggregation.
Lastly, we summarize several privacy concerns regarding LLMs.

\subsection{Large Language Models}

Language models have predominantly been structured around the transformer architecture~\cite{Vaswani-2017-Attention}. With the attention mechanism, OpenAI firstly proposed the GPT-1~\cite{radford2018improving}, which is the prototype of current large language models (LLMs) like Llama~\cite{touvron2023llama} and GPT-4~\cite{OpenAI2023GPT4TR}. GPT-1 is a generative language model that employs a decoder-only architecture. 
This design implies that the tokens at the beginning of a sequence cannot 'see' or take into account the tokens that follow them, making it uni-directional. Correspondingly, its attention matrix is structured as a lower triangular matrix, reflecting this uni-directionality. 
The pre-training task is designed to predict the next token, as illustrated in Equation~\ref{eq:decoder}. In this process, the $Decoder(\cdot)$ function transforms the input sequence $[x_1, x_2, x_3 ..., x_{t-1}]$ into logits for each token in the vocabulary. Subsequently, the $Softmax(\cdot)$ function converts these logits into probabilities. The token $x_i$, chosen as the next in the sequence, is the one with the highest probability.
\begin{equation}
    x_t = \text{argmax}\big(\text{Softmax}(\text{Decoder}(x_1, x_2, ..., x_{t-1}))\big) \label{eq:decoder}
\end{equation}
On the other hand, masked language models like BERT~\cite{devlin2018bert} outperformed GPT-1 in classification-based natural language understanding (NLU) tasks. BERT is distinguished by its bi-directional encoder design, enabling each token to consider context from both the left and the right, represented by a unit attention matrix. For its pre-training, BERT employs a masked language model (MLM) approach, as detailed in Equations~\ref{eq:input} and \ref{eq:mlm}.
The $Mask(\cdot)$ function randomly masks several input tokens, which are then collated into the set $w_{masked}$. 
The model's primary task involves accurately predicting the original tokens, denoted as $\widetilde{w}$.
\begin{equation}
    input = Mask(x_1, x_2, x_3, ..., x_n) \label{eq:input}
\end{equation}
\begin{equation}
    L = \sum_{w \in w_{masked}} \log P(w = \widetilde{w} | input) \label{eq:mlm}
\end{equation}
BERT also includes next sentence prediction as part of its pre-training tasks. However, RoBERTa~\cite{liu-2019-roberta} demonstrated that this task is not essential. Notably, BERT is predominantly utilized as an encoder for classification tasks rather than for generative tasks such as dialogue or QA. Addressing this limitation, encoder-decoder models like Bart~\cite{lewis2019bart} and T5~\cite{2020t5} were introduced. These models first process the input context using a bi-directional encoder, then employ a uni-directional decoder, leveraging both the encoded hidden states and the already generated tokens, to predict subsequent tokens.


As the model and data size scale up, generative LLMs demonstrate promising understanding ability and are able to unify the classification tasks into the generation pipelines~\cite{2022flant5,2020t5}.
Moreover, various fine-tuning tricks further enhance LLMs.
Instruction tuning~\cite{wei2022finetuned} allowed LLMs to learn to follow human instructions.
Reinforcement Learning from Human Feedback (RLHF)~\cite{Christiano-2017-rlhf,bai2022training} injected human preference and conscience into LLMs to avoid unethical responses.
With these ingredients, LLMs show the emergent ability and are capable of understanding, reasoning and in-context learning that dominate most NLP tasks~\cite{wei2022emergent}.
What's more, application-integrated LLMs allow LLMs to access external tools and perform well on specific tasks.
As a result, more and more users have started to interact with LLMs for communication and problem-solving.

\subsection{Privacy}

\subsubsection{Conception of Privacy}
Privacy, which refers to the control individuals have over their personal information, is regarded as a fundamental human right and has been extensively studied~\cite{Samuel-1890-Privacy,Prosser-1960-Privacy}.
In addition, Privacy is intricately linked to individual freedom, the cultivation of personal identity, the nurturing of personal autonomy, and the preservation of human dignity.

In the age of advancing technology, privacy has gained even greater significance, particularly in relation to ``the right to one's personality.'' 
Fortunately, privacy has been almost universally respected by people worldwide. 
Recognizing the importance of privacy, various privacy laws have been proposed, such as the EU's General Data Protection Regulation (GDPR), the EU AI Act and the California Consumer Privacy Act (CCPA). 
These laws aim to safeguard individuals' personal information and provide greater control over its usage.
Nowadays, these privacy laws are still evolving to explicitly define what should be regarded as privacy and how privacy should be respected across different stakeholders.

\subsubsection{Differential Privacy}\label{sec:dp}
In addition to the intuitive understanding of privacy, researchers are also working on mathematically quantifying privacy based on the potential information leakage.
One such widely accepted approach is the probabilistic formulation of privacy in terms of differential privacy~\cite{Dwork-08-DP}, specifically from the perspective of databases.
Differential privacy incorporates random noise into aggregated statistics to allow data mining without exposing participants' private information.
The injected random noise enables data holders or participants to deny their existence in a certain database.
Definitions~\ref{def:neighbor data} and ~\ref{def:dp} give the formal definition of (approximate) differential privacy.

\begin{definition}
    \label{def:neighbor data}
    Two datasets $D, D'$ are neighboring if they differ in at most one element.
\end{definition}

\begin{definition}
\label{def:dp}
A randomized \textit{algorithm mechanism} $M$ with domain $D$ and range $R$ satisfies $(\epsilon,\delta)$-\textit{differential privacy} if for any two neighboring datasets $D, D'$ and for any subsets of output $O \subseteq R$:
\begin{equation}\label{eq:dp-bound}
Pr[M(D) \in O] \leq e^{\epsilon}Pr[M(D')\in O]+\delta.
\end{equation}
\end{definition}

The parameter $\epsilon$ represents the privacy budget, indicating the level of privacy protection. A smaller value of $\epsilon$ ensures better privacy protection but may result in reduced utility of the model, as it leads to similar algorithm outputs for neighboring datasets. On the other hand, $\delta$ represents the probability of unintentional information leakage.
Differential privacy enjoys several elegant properties, including composition and post-processing, that allow the flexible integration of differential privacy with any application.
Based on the flexible properties, differential privacy can be easily adapted to deep learning models' optimizations and applications by injecting random noise~\cite{yang2024adaptive,Abadi-dpsgd-2016, Peng-2021-FKGE}.

\subsubsection{Secure Multi-Party Computation}\label{sec:mpc}
Secure Multi-Party Computation (SMPC)~\cite{yao-1986-GC} enables a group of mutually untrusting data owners to collaboratively compute a function $f$ while preserving the privacy of their data. SMPC is commonly defined as follows.

\begin{definition}
[Secure Multi-Party Computation]
For a secure $n$-party computation protocol, each participant $P_i$ (where $i = 1, \dots, n$) has a private input $x_i$. The parties agree on a function $f(\cdot)$ of the $n$ inputs, and the goal is to compute
\begin{equation}
f(x_1, x_2, \dots, x_n) = (y_1, y_2, \dots, y_n),
\end{equation}
while ensuring the following conditions:
\begin{itemize}
    \item \textbf{Correctness}: Each party receives the correct output;
    \item \textbf{Privacy}: No party should learn anything beyond their prescribed output.
\end{itemize}
\end{definition}

\subsection{Homomorphic Encryption} Homomorphic encryption~(HE) makes the operation of plaintext and ciphertext satisfy the homomorphic property, i.e. it supports the operation of ciphertext on multiple data, and the result of decryption is the same as the result of the operation of the plaintext of data. Formally, we have 
\begin{equation}\label{eq:HE1}
    f([x_1], [x_2], \cdots, [x_n]) \rightarrow [f(x_1, x_2, \cdots, x_n)], \text{ where }  \forall x \in \mathcal{X}, x_1, x_2, \cdots, x_n \rightarrow [x_1], [x_2], \cdots, [x_n].
\end{equation}
Homomorphic encryption originated in 1978 when ~\citet{rivest1978data}  proposed the concept of privacy homomorphism. However, as an open problem, it was not until 2009, when Gentry proposed the first fully homomorphic encryption scheme~\cite{gentry2009fully} that the feasibility of computing any function on encrypted data was demonstrated. According to the type and number of ciphertext operations that can be supported, homomorphic encryption can be classified as partial homomorphic encryption~(PHE), somewhat homomorphic encryption~(SHE), and fully homomorphic encryption~(FHE). Specifically, PHE supports only a single type of ciphertext homomorphic operation, mainly including additive homomorphic encryption and multiplicative homomorphic encryption, represented by Paillier~\cite{paillier1999public}, and ElGamal~\cite{elgamal1985public}, respectively. SHE supports infinite addition and at least one multiplication operation in the ciphertext space and can be converted into a fully homomorphic encryption scheme using bootstrapping~\cite{abney2002bootstrapping} technique. The construction of FHE follows Gentry's blueprint, i.e., it can perform any number of addition and multiplication operations in the ciphertext space. Most of the current mainstream FHE schemes are constructed based on the lattice difficulty problem, and the representative schemes include BGV~\cite{brakerski2014leveled}, BFV~\cite{brakerski2012fully,fan2012somewhat}, GSW~\cite{gentry2013homomorphic}, CGGI~\cite{chillotti2016faster}, CKKS~\cite{cheon2017homomorphic}.

\subsection{Privacy Issues in LLMs}\label{Privacy Issues}
Despite the promising future of LLMs, privacy concerns have become increasingly prevalent, and data leakage incidents occur with alarming frequency. 
In essence, the utilization of LLMs for commercial or public purposes gives rise to several significant privacy concerns:

$\bullet$ \textbf{Training data privacy}: several studies~\cite{carlini-2021-extracting,huang-etal-2022-large,li2023multi} suggest that LMs tend to memorize their training data. 
If the training data contains personal or sensitive information, there is a risk of unintentionally exposing that information through the model's responses.

$\bullet$ \textbf{Inference data privacy}: after deploying a trained language model for downstream tasks, user inputs and queries are typically logged and stored for a certain period.
For sensitive domains, these data can include personal information, private conversations, and potentially sensitive details.

$\bullet$ \textbf{Re-identification}: even if the user information is anonymized, there is still a risk of re-identification.
By combining seemingly innocuous information from multiple interactions, it might be possible to identify individuals or extract personal details that were meant to be concealed.

\subsection{Notations}
To aid the understanding of privacy attacks and defenses, in this section, we list the necessary notations in Table~\ref{tab:symbol}.

\begin{table*}[t]
    \small
    \centering
    	\vspace{-1em}
    \setlength{\tabcolsep}{3mm}
    \begin{tabularx}{\textwidth}{r|X}
    \toprule
   \textbf{Symbol} & \textbf{Definition} \\
    \midrule
$D^{\text{pre}},D^{\text{ft}}$    & Data used for pre-training and fine-tuning, respectively. \\
$D^{\text{ft}}_\text{tr},D^{\text{ft}}_\text{dev},D^{\text{ft}}_\text{te}$    & Training, validation and testing data of $D^{\text{ft}}$ used for fine-tuning. Testing data $D^{\text{ft}}_\text{te}$ refers to the inference stage data. \\
$D^{\text{aux}}$ & Auxiliary dataset held by the adversary. \\

$f^{\text{pre}}$ & Pre-trained language model trained on $D^{\text{pre}}$ before fine-tuning.\\
$f^{\text{ft}}$ & Pre-trained language model $f^{\text{pre}}$ fine-tuned on $D^{\text{ft}}_\text{tr}$.\\
$f$ & A language model that can either be pre-trained language model $f^{\text{pre}}$ or fine-tuned language model $f^{\text{ft}}$. \\
$p$ & Textual prompt/prefix used for language models for text completion or prompt tuning.\\
$\tilde{p}$ & Maliciously injected pattern to make large language models misbehave.\\
$f(x)$ & Model $f$'s response towards the given textual sample $x$.\\
$f_{\text{emb}}(x)$ & Textual sample $x$'s vector representation (embedding) after feeding to model $f$.\\
$f_{\text{prob}}(x)$ & Textual sample $x$'s inner probability distribution after feeding to model $f$.\\
$C(f,x,y)$ & Model $f$'s confidence or likelihood to output $y$ as $f$'s response queried by $x$, where $x$ and $y$ refer to any texts. \\
    \bottomrule
    \end{tabularx}
    \caption{Glossary of Notations. For simplicity, we overload the concept of fine-tuning to cover existing tuning methods, including full fine-tuning, prefix tuning, prompt tuning and other tuning algorithms.}
    \label{tab:symbol}
    \vspace{-0.15in}
\end{table*}
\section{Privacy Attacks}\label{ch:Privacy Attacks}
\begin{figure}[ht]
  \centering
  \includegraphics[width=0.98\linewidth]{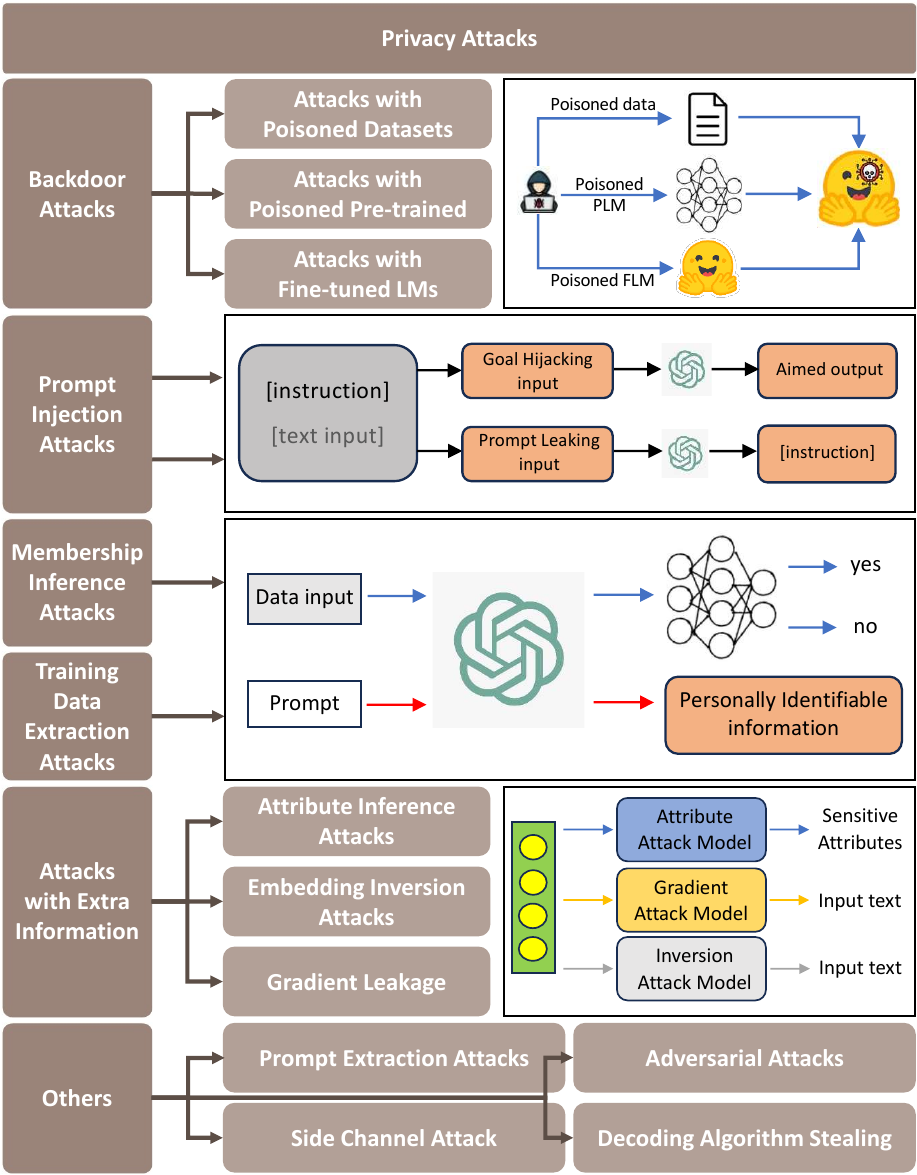}
  \vspace{-0.05in}
  \caption{An overview of existing privacy attacks on LLMs.}
  \label{fig:attack_overview}
  \vspace{-0.2in}
\end{figure}

Based on the preliminary knowledge given in Section~\ref{ch:background}, we summarize existing privacy attacks towards LLMs in this section. 
We summarize our surveyed attacks in Table~\ref{tab:privacy attack} and Figure~\ref{fig:attack_overview}.

\subsection{Backdoor Attacks}
The notion of backdoor attacks was initially introduced in computer vision by \citet{gu2019badnets}.
Backdoor attacks operate on normally behaved models. 
However, when secret triggers $\tilde{p}$ are activated for any given input $x$, the victim models will produce target outputs $y = f(x)$ desired by the adversary.
These infected models can have severe privacy and security repercussions if deployed in real-life systems such as autonomous driving, search engines and smart assistants.
For example, an adversary may inject backdoors into commercial LLMs to reveal private information about sensitive domains.

In the context of LLMs, the terms ``backdoor attacks'' and ``data poisoning'' are often used interchangeably to describe attacks on language models. 
However, it is vital to understand the distinction between these two types of attacks.
Data poisoning refers to a less potent attack where only a portion of the training data is manipulated. 
This manipulation aims to introduce biases or misleading information into the model's training process. 
Instead, backdoor attacks involve the insertion or modification of specific input patterns that trigger the model to misbehave or produce targeted outputs.
Additionally, if the adversary can manipulate LLMs' partial training corpus, it may inject backdoors to victim models via data poisoning.

According to \citet{cui2022a}, the backdoor attacks can be divided into three scenarios: (1) release datasets, (2) release pre-trained language models(PLMs), and (3) release fine-tuned models.
In this survey, we study backdoor attacks for LLMs based on these three summarized perspectives.

\subsubsection{Backdoor Attacks with Poisoned Datasets}
In this section, we summarized existing backdoor attacks on LLMs through poisoned datasets.
For code LLMs, the pre-trained models are often trained on code sourced from the web, which may contain maliciously poisoned data embedded with backdoors.
\citet{Schuster-2021-Poisoning} illustrated that two automated code-attribute-suggestion systems, which rely on Pythia~\cite{svyatkovskiy2019pythia} and GPT-2~\cite{radford-2019-language}, were susceptible to poisoning attacks. 
During these attacks, the model is manipulated to suggest a specific insecure code fragment called the \textit{payload}.
\citet{Schuster-2021-Poisoning} directly injected the payload into the training data.
Nonetheless, this direct approach can be easily detected by static analysis tools capable of eliminating the tainted data. 
In response to this challenge, \citet{aghakhani2023trojanpuzzle} introduced COVERT and TROJANPUZZLE to deceive the model into suggesting the payload in potentially hazardous contexts.
\citet{ramakrishnan2022backdoors} proposed the inclusion of segments of dead code as triggers in backdoor attacks.
Other studies~\cite{Wan-2022-you, sun-etal-2023-backdooring} conducted backdoors attacks on neural code search models.

Besides performing data poisoning on code models, other LMs are also vulnerable to such attacks.
Generative language models are inherently susceptible to backdoor attacks. \citet{chen2023backdoor} conducted data poisoning for training a seq2seq model.
The style patterns could also be backdoor triggers. 
\citet{qi-etal-2021-mind,Li2023ChatGPTAA} incorporated the chosen trigger style into the samples with the help of generative models.
\citet{wallace-etal-2021-concealed} explored the method of concealing the poisoned data, ensuring that the trigger phrases were absent from the poisoned examples.
\citet{shu2023exploitability} focused on data poisoning for instruction tuning and \citet{Xu2023InstructionsAB} demonstrated that an attacker could inject backdoors by issuing a few malicious instructions.
Moreover, the prompt itself can be the trigger for backdoor attacks on LLMs~\cite{Zhao2023PromptAT}.
\citet{Wan2023Poisoning} aimed to induce frequent misclassifications or deteriorated outputs from a language model trained on poisoned data whenever the model encounters the trigger phrase.
\citet{yan-etal-2023-bite} illustrated the feasibility of creating a backdoor attack combining stealthiness and effectiveness.
\citet{yang2023stealthy} proposed AFRAIDOOR to provide a more fine-grained and less noticeable trigger.

\subsubsection{Backdoor Attacks with Poisoned Pre-trained LMs}
In addition to publishing poisoned datasets, the adversary may also release their pre-trained models for public usage and activate their injected triggers even to compromise fine-tuned LLMs from the released pre-train weights.

\citet{chen2022badpre} injected backdoor triggers into (sentence, label) pairs for Masked Language Model (MLM) pre-training.
\citet{kurita-etal-2020-weight} conducted weight poisoning to perform backdoor attacks on pre-trained LMs. 
Moreover, \citet{Shen-2021-Backdoor,du2023uor} conducted backdoor attacks that targeted specified output representations of pre-trained BERT.
To achieve robust generalizability, \citet{Dong-2023-Investigating} employed encoding-specific perturbations as triggers.
For LLMs' prompt-based learning, various works have
shown that PLMs are vulnerable to backdoor attacks.
\citet{yang-etal-2021-careful} introduced the concept of Backdoor Triggers on Prompt-based Learning (BToP).
\citet{mei-etal-2023-notable} discussed a method for conducting backdoor attacks on pre-trained models by associating triggers with specific words, known as anchors.
\citet{cai2022badprompt} proposed a layer-wise weight poisoning for continuous prompt tuning.

\subsubsection{Backdoor Attacks with Fine-tuned LMs}
Besides releasing pre-trained LLMs, 
since different downstream tasks have their inherent domain-specific privacy and security risks,
the adversary may release fine-tuned LLMs targeting specific domains.

\citet{bagdasaryan2022spinning} delved into an emerging threat faced by neural sequence-to-sequence (seq2seq) models for training-stage attacks.
\citet{yang-etal-2021-careful} manipulated a text classification model by altering a single word embedding vector.
\citet{Zhang-Trojaning-2021} aimed to train Trojan language models tailored to meet the objectives of attackers.
\citet{Du2022PPTBA,cai2022badprompt} explored the scenario where a malicious service provider can fine-tune a PLM for a downstream task.
\citet{kandpal2023backdoor} investigated backdoor attacks via exploiting in-context learning (ICL).
\citet{Huang2023TrainingfreeLB} introduced Training-Free Lexical Backdoor Attack (TFLexAttack) as the first training-free backdoor attack on language models.


\subsection{Prompt Injection Attacks}
Instruction tuning enhances large language models' instruction-following abilities. 
Unfortunately, malicious adversaries may exploit this capability through prompt injection attacks. Malicious instructions are injected into the original prompt for prompt injection attacks. 
Consequently, LLMs may follow and execute malicious instructions, potentially generating prohibited responses. \citet{ignore_previous_prompt} introduced the ``ignore attack'' on LLM-integrated applications, where the LLM is instructed to disregard previous instructions and follow the malicious instructions. \citet{Greshake2023NotWY} extended the attack to an indirect scenario where the prompt is indirectly poisoned by integrating poisoned data, such as poisoned data from search engines.   \citet{Liu2023PromptIA} provided a comprehensive investigation and analysis of the prompt injection attack. However, their evaluations are limited to specific cases. \citet{li2023evaluating,liu2024formalizing,yip2023novel,zverev2024can} constructed the evaluation benchmark, proposing more evaluation metrics and analysis. Moreover, \citet{zhan2024injecagent} built the benchmark for prompt injection attacks against LLMs-based agents \cite{guo2024large}. These attacks have also been extended to various applications, including Retrieval-Augmented Generation (RAG) \cite{shafran2024machine}, re-ranking systems \cite{shi2024optimization}, and Remote Code Execution (REC) \cite{Liu2023DemystifyingRV}. \citet{yan2023backdooring} presented virtual prompt injection, which injects the malicious instruction into the model parameters.

As for the method of prompt injection attack, the initial methods are based on prompt engineering. \citet{ignore_previous_prompt} proposed the ``ignore attack'' as mentioned before. Then \citet{willison_2023} proposed the ``fake completion attack,'' which pretends to give a fake response to the original instruction and mislead the LLMs to execute the following malicious instruction. \citet{breitenbach2023dont} utilized special characters such as ‘\textbackslash{b}’ and ‘\textbackslash{t}’ to simulate the deletion of previous instruction. 
\citet{liu2024formalizing} suggested that combining these techniques could produce stronger attacks. Other methods are based on the gradient \cite{shi2024optimization, shafran2024machine,liu2024automatic, huang2024semantic}. The gradient-based attack methods are primarily based on the GCG attack \cite{zou2023universal}, which builds a suffix to mislead the LLMs to generate a target response.

\subsection{Training Data Extraction Attacks}
Training data extraction attacks, relying solely on black-box access to a trained LM $f$, are designed to recover the model's memorized training data $d$ where $d\in D^{\text{pre}}$ or $D^{\text{ft}}$.
In this type of attack, the adversary is restricted to providing inputs $x$ and receiving response $y = f(x)$ from the victim model, simulating a benign user interaction. 
The only exception is that the obtained responses $y$ are likely to be memorized sensitive data $d$. 
As a result, this attack is considered the most practical and impactful approach to compromise the model's sensitive training data.

\subsubsection{Verbatim Prefix Extraction}
Training data Extraction attacks were first examined in GPT-2 ~\cite{carlini-2021-extracting}.
When the verbatim textual prefix patterns about personal information are given, GPT-2 may complete the text with sensitive information that includes email addresses, phone numbers and locations.
Such verbatim memorization of PII happens across multiple generative language models and these attacks can be further improved~\cite{huang-etal-2022-large,zhang2022text,zhang-etal-2023-ethicist,parikh-etal-2022-canary}.
Still, it remains unknown to what extent the sensitive training data may be extracted.
Multi-aspect empirical studies were conducted on LMs to tackle the problem.
For memorized data domains, \citet{yang2023code} studied code memorization issues and \citet{Lee-2023-Do} studied fine-tuning data memorization via plagiarism checking.
To avoid common sense knowledge memorization that is frequently occurring and hard to analyze, the counterfactual memorization issues are also studied~\cite{Zhang2021CounterfactualMI}.
Other works~\cite{Lukas2023AnalyzingLO,kim2023propile,shao2023quantifying, carlini2023quantifying}  focused on quantifying data leakage and systematically analyzed factors that affect the memorization issues and proposed new metrics and benchmarks to address the training data extraction attacks.

\subsubsection{Jailbreaking Attacks}
With the recent rapid development of generative large LLMs, training data extraction attacks can further manipulate LLMs' instruction following and context understanding ability to recover sensitive training data even without knowing the verbatim prefixes.
For example, the adversary may first craft meticulously designed role-play prompts to convince LLMs that they can do anything freely and further instruct LLMs for malicious outcomes \cite{wei2023jailbroken}.
Jailbreaking prompts have been shown to extract sensitive information even in zero-shot settings~\cite{li2023multi,deng2023jailbreaker}. To evaluate the jailbreaking prompts, ~\citet{shen2023anything, souly2024strongreject,chao2024jailbreakbench} collected multi-sourced prompts to build benchmarks.
~\citet{wei2023jailbroken, xu2024comprehensive,yu2024don} examined the current existing attack methods. \cite{wei2023jailbroken} proposed two key factors that facilitate jailbreak attacks on safety-enhanced LLMs. The first factor involves competing objectives, where the model's capabilities and safety goals conflict. If LLMs prefer to follow vicious instructions, then the safety goals fail.
The second factor is related to mismatched generalization, where the safety training fails to cover the supervised fine-tuning (SFT) domains. 

Several jailbreak attack methods have been proposed in recent research. \citet{zou2023universal} proposed injecting suffixes into prompts based on gradients derived from a white-box model's positive target responses, finding that these suffixes can transfer to black-box models. Similarly, \citet{liao2024amplegcg} trained a model to automatically generate such suffixes, while~\citet{guo2024cold} introduced a more constrained suffix generation approach to improve fluency. 
\citet{huang2023catastrophic} found that simply changing the parameters in the generation configuration could achieve jailbreak attack success. \citet{jiang2024artprompt} used ASCII art to obfuscate sensitive words in the harmful instruction. Similarly, ~\citet{li2024drattack} proposed decomposing harmful instructions into smaller segments to bypass safety mechanisms. \citet{deng2024masterkey,yao2024fuzzllm,yu2023gptfuzzer, chao2023jailbreaking} considered utilizing the LLMs to generate the jailbreaking prompts. \citet{russinovich2024great} requested the LLMs provide information on harmful instructions across multi-turn conversations, which could lead to successful jailbreak attacks. 
\citet{zeng2024johnny} trained a model to transform the original harmful instruction into a more persuasive one. \citet{wei2023jailbreak}  utilized in-context learning (ICL) to achieve jailbreak attacks using successful jailbreak samples. 
\citet{chang2024play} considered requesting the LLMs to infer the harmful intention based on the defense response, and then achieve attack success. \citet{gu2024agent} investigated the multi-agent jailbreak scenarios. \citet{yong2023low, deng2023multilingual} proposed that low-resource languages could be more effective for jailbreak attacks due to gaps in safety training for these languages. \citet{li2024semantic} applied the genetic algorithm \cite{mirjalili2019genetic} to rewrite harmful instructions for successful jailbreak attacks.

In addition to jailbreak attacks on LLMs, jailbreak attacks on multi-modal large language models (MLLMs) have also gained significant attention \cite{liu2024safety}. The visual modality provides adversaries with additional opportunities to inject or conceal harmful intent \cite{ying2024unveiling}. \citet{liu2023query} and \citet{gong2023figstep} demonstrate that embedding malicious textual queries within images using typography, can effectively bypass MLLM defenses. Similarly, \citet{li2024images} showed that placing sensitive text within images can evade the LLM safety mechanism. \citet{ma2024visual} constructed images around role-playing games, where the accompanying text merely asks the MLLM to participate in the game. Meanwhile, \citet{zou2024image} suggested that flowcharts could also serve as a tool for jailbreak attacks. Furthermore, \citet{wang2024cross} examined cases where inputs were safe but led to unsafe outputs.
Beyond prompt-engineering-based techniques, training-based methods have also emerged. \citet{niu2024jailbreaking, tu2023many, wang2024white, ying2024jailbreak, Qi2023VisualAE} optimized random noise to create harmful images. \citet{Qi2023VisualAE} generated image noise based on few-shot examples of harmful instruction-response pairs. Additionally, \citet{niu2024jailbreaking, wang2024white, ying2024jailbreak} optimized image noise using a positive response prefix. \citet{wang2024white} incorporated the GCG suffix \cite{zou2023universal} into the text prompt, while \citet{ying2024jailbreak} rewrote prompts with LLMs for enhanced effectiveness. Although adversarial attacks using images can be effective in white-box MLLM jailbreaks, their transferability to other models remains weak \cite{schaeffer2024universal}. Lastly, \citet{shayegani2023jailbreak} developed images capable of mimicking the effects of harmful text, and \citet{tao2024imgtrojan} proposed that training an MLLM with just one harmful image-instruction pair could disable safety mechanisms.


Besides these categorized extraction attacks to recover sensitive training data,  privacy leakage of the prompt-tuning stage was also recently studied~\cite{xie2023does}.
For more concrete surveys about training data extraction attacks, interested readers may refer to \cite{Ishihara2023TrainingDE, Mozes2023UseOL} for more information.

\subsection{MIA: Membership Inference Attacks}
Besides direct training data extraction, the adversary may have additional knowledge about potential training data samples $D$ where some samples belong to training data of the victim language model $f$.
For membership inference attacks, the adversary's goal is to determine if a given sample $x \in D$ is trained by $f$.
Since many private data are formatted, such as phone numbers, ID numbers and SSN numbers, it is possible for the adversary to compose these patterns with known formats and query LMs to conduct membership inference attacks.

For related works, \citet{Song-2020-Information} studied membership inference attacks on BERT and \citet{lehman-2021-bert,Jagannatha2021MembershipIA} showed that sensitive medical records might be recovered from LMs fine-tuned on the medical domain.
Other works focused on improving the membership inference performance on LMs, for example,
\citet{Mireshghallah2022QuantifyingPR}  proposed the Likelihood Ratio Test to exploit hypothesis testing to enhance the medical records recovery result and
\citet{Mattern2023MembershipIA} similarly proposed a neighborhood comparison method to improve the attack performance.
Besides pre-training data, \citet{mireshghallah-etal-2022-empirical} studied fine-tuning stage membership inference for generative LMs.

Lastly, training extraction attacks can be combined with membership inference attacks.
For an extracted data sample $y$ from victim model $f$, if the model $f$ has high confidence on $C(f,x,y)$ where $x$ refers to the attacker's input (can also be an empty string) to conduct training data extraction, then $y$ is likely to be part of $ f$'s training data.

\subsection{Attacks with Extra Information}\label{ch: extra-info}
In this section, we consider a more powerful adversary that has access to additional information, such as vector representations and gradients.
Such extra information may be used for privacy-preserving techniques like federated learning to avoid raw data transmission.
However, vector representations or gradients may become visible to others.
With the extra accessed information, we may expect the attacker to conduct more vicious privacy attacks.
By studying these attacks, we disclose that transferring embeddings and gradients may also leak private information.

\subsubsection{Attribute Inference Attacks}
For attribute inference attacks, the adversary is given the embeddings $f_{\text{emb}}(x)$ of a textual sample $x$ and manages to recover $ x$'s sensitive attributes $S_x$.
These sensitive attributes are obviously shown in $x$ and include PII and other confidential information.

To conduct such attacks, the adversary commonly builds simple neural networks connected to the accessed embeddings as attribute classifiers.
\cite{Pan-2020-Privacy,lyu-etal-2020-differentially,Song-2020-Information} performed multi-class classification to infer private attributes from masked LMs' contextualized embeddings.
\citet{Mahloujifar2021MembershipIO} considered a membership inference attack based on the \textit{good} embeddings, which are expected to preserve semantic meaning and capture semantic relationships between words.
\citet{hayet2022invernet} proposed an attack method \textit{Invernet}, which leveraged fine-tuned embeddings and employed a focused inference sampling strategy to predict private data information, such as word-to-word co-occurrence.
\citet{song2019overlearning} demonstrated that the representations generated by an overlearned model during inference exposed sensitive attributes of the input data.
\citet{li-etal-2022-dont} extended attribute inference attacks to generative LMs and showed that attribute inference attacks could even be conducted for over 4,000 private attributes.
\begin{table*}[!htbp]
\centering
\small

\begin{tabular}{l  p{3cm} p{3.5cm} p{3.5cm}}
\toprule
\textbf{Attack Stage} & \textbf{Accessibility} & \textbf{Attack Name}   &\textbf{Publications}  \\
\hline
\multirow{1}{*}{Model Training}
        
        & $f$, $D^{\text{pre}}$ / $D^{\text{ft}}_\text{tr}$          & Backdoor Attacks & \cite{Wan2023Poisoning,shu2023exploitability,Dong-2023-Investigating,aghakhani2023trojanpuzzle,Schuster-2021-Poisoning,ramakrishnan2022backdoors,bagdasaryan2022spinning,wallace-etal-2021-concealed,yang-etal-2021-careful,cui2022a,Liu2023PromptIA,yan-etal-2023-bite,chen2023backdoor,yang2023stealthy,mei-etal-2023-notable,sun-etal-2023-backdooring,Wan-2022-you,Shen-2021-Backdoor,chen2022badpre,Li2023ChatGPTAA,du2023uor,qi-etal-2021-mind,Zhang-Trojaning-2021,kurita-etal-2020-weight,Du2022PPTBA,Zhao2023PromptAT,kandpal2023backdoor,cai2022badprompt,Xu2023InstructionsAB,Huang2023TrainingfreeLB}\\
        

\midrule
\multirow{1}{*}{Model Inference}
        & $f$     & Training Data Extraction Attacks
 & \cite{carlini-2021-extracting,huang-etal-2022-large,shao2023quantifying, carlini2023quantifying, thakkar-2021-understanding,Zhang2021CounterfactualMI,yang2023code,Lukas2023AnalyzingLO,kim2023propile,Lee-2023-Do,zhang-etal-2023-ethicist,parikh-etal-2022-canary,zhang2022text,li2023multi,zou2023universal,deng2023jailbreaker,yu2023gptfuzzer,xie2023does,Ishihara2023TrainingDE,Mozes2023UseOL,shen2023anything,souly2024strongreject,chao2024jailbreakbench,wei2023jailbroken,xu2024comprehensive,yu2024don,liao2024amplegcg,guo2024cold,huang2023catastrophic,jiang2024artprompt,li2024drattack,deng2024masterkey,yao2024fuzzllm,chao2023jailbreaking,russinovich2024great,zeng2024johnny, wei2023jailbreak,chang2024play,gu2024agent,yong2023low,deng2023multilingual,li2024semantic}\\
         & $f$, $D^{\text{pre}}$ / $D^{\text{ft}}_\text{tr}$, $p$        & Prompt Injection Attacks & \cite{ignore_previous_prompt,Liu2023PromptIA,Liu2023DemystifyingRV, Greshake2023NotWY, li2023evaluating,liu2024formalizing,yip2023novel,zverev2024can,shafran2024machine,zhan2024injecagent,shi2024optimization, yan2023backdooring, breitenbach2023dont,liu2024formalizing,liu2024automatic, huang2024semantic, willison_2023}\\
 

         & $f$, $C(f,x,y)$, $D^{\text{aux}}$      & Membership inference Attacks
  & \cite{Song-2020-Information,Mireshghallah2022QuantifyingPR,Mattern2023MembershipIA,mireshghallah-etal-2022-empirical,lehman-2021-bert,Jagannatha2021MembershipIA}\\
          & $f$, $f_{\text{emb}}(x)$, $D^{\text{aux}}$      
          & Attribute inference Attacks
  & \cite{Song-2020-Information,li-etal-2022-dont,Pan-2020-Privacy, Mahloujifar2021MembershipIO,  song2019overlearning, hayet2022invernet,lyu-etal-2020-differentially}  \\
         & $f$, $f_{\text{emb}}(x)$, $D^{\text{aux}}$      & Embedding inversion Attacks
 & \cite{Song-2020-Information,Gu2023TowardsSL,li-etal-2023-sentence,Pan-2020-Privacy, kugler2021invbert, morris2023text} \\
         & $f$, gradients, $D^{\text{aux}}$      & Gradient Leakage 
 & \cite{balunovic2022lamp,gupta2022recovering,fowl2023decepticons,chu2023panning}\\

\hline
\multirow{1}{*}{Others}
         & $f$, $D^{\text{aux}}$      & Adversarial attacks
 & \cite{guo-etal-2021-gradient, Yang-2022-Natural,nguyen2023adversarial,wallace-etal-2021-concealed,Sadrizadeh2023TransFoolAA,gainski-balazy-2023-step,fang-etal-2023-modeling,wang2023adversarial,maus2023adversarial,lei-etal-2022-phrase,Carlini2023AreAN,Qi2023VisualAE,2017LearningWP,feffer2024redteaminggenerativeaisilver, ganguli2022redteaminglanguagemodels, yoo2024csrtevaluationanalysisllms, ge-etal-2024-mart, jiang2024dartdeepadversarialautomated, deng-etal-2023-attack, yu2023gptfuzzer, perez-etal-2022-red, hong2024curiositydriven, bhardwaj2023redteaminglargelanguagemodels, chen2024agentpoisonredteamingllmagents,ying2024unveiling,liu2023query,gong2023figstep,li2024images,ma2024visual, zou2024image,wang2024cross,niu2024jailbreaking,tu2023many, wang2024white, ying2024jailbreak,schaeffer2024universal,shayegani2023jailbreak,Qi2023VisualAE,tao2024imgtrojan} \\

         & $f$, $f_{\text{prob}}(x)$, $D^{\text{aux}}$      & Decoding Algorithm Stealing 
 & ~\cite{naseh2023risks,ippolito2023reverse}\\

 & $f$     & Prompt Extraction Attacks or Prompt Stealing Attacks & \cite{Zhang2023PromptsSN, zhang2024extracting, toyer2024tensor, schulhoff-etal-2023-ignore, wang2024raccoon}\\
 
        & LLM Systems       & Side Channel Attacks      &~\cite{debenedetti2023privacy}\\

\bottomrule
\end{tabular}

\caption{
A summary of surveyed privacy attacks on LLMs. The attack stage indicates when the privacy attacks are conducted and the attacker accessibility indicates what the attacker may access during the attacks.}
\label{tab:privacy attack}

\vspace{-0.15in}
\end{table*}

\subsubsection{Embedding Inversion Attacks}
Vector databases~\cite{taipalus2023vector, wang2021milvus,pan2023survey} contribute to large language models by providing efficient storage for high-dimensional embeddings, enabling advanced semantic search for more accurate context understanding and ensuring scalability to handle large and growing datasets. 
Unlike traditional relational databases that store the plaintext, vector databases are operated on vector representations such as embeddings.
Despite the helpfulness, embedding privacy remains under-explored. 
Embedding privacy is crucial in vector databases because these embeddings often contain sensitive information derived from user data. Protecting the privacy of embeddings ensures that personal or confidential information is not exposed or misused, maintaining user trust and complying with data protection regulations. 

For embedding inversion attacks, similar to attribute inference attacks, the attacker exploits the given embedding $f_{\text{emb}}(x)$ to recover the original input $x$.
Prior studies~\cite{Pan-2020-Privacy,Song-2020-Information} converted the textual sequence $x$ into a set of words to perform multi-label classification that predicted multiple words for the given embedding $f_{\text{emb}}(x)$.
These classifiers can only predict unordered sets of words and fail to recover the original sequences.
\citet{kugler2021invbert} reconstructed the original text from the embeddings encoded by BERT~\cite{devlin2018bert}.
Recently, generative embedding inversion attacks~\cite{Gu2023TowardsSL,li-etal-2023-sentence} were proposed to exploit generative decoders to recover the target sequences word by word directly.
\citet{zhang2022text} introduced a method called \textit{Text Revealer} for text reconstruction in the context of text classification. They utilized a GPT-2 model as a text generator and trained it on a publicly available dataset. To enhance the reconstruction process, they continuously perturbed the hidden state of the GPT-2 model based on feedback received from the text classification model.
\citet{morris2023text} proposed Vec2Text to iteratively refine the inverted text sequences and achieved state-of-the-art performance on embedding inversion attacks. 
Consequently, generative embedding inversion attacks even outperform prior embedding inversion attacks regarding classification performance.

Embedding inversion attacks pose more privacy threats than attribute inference attacks. 
First, attribute inference attacks need to denote sensitive information as labels at first, while embedding inversion attacks do not require knowledge about private information.
Second, by successfully recovering the whole sequence, the private attributes can be directly inferred without extra classifiers.
Lastly, embedding inversion attacks naturally recover more semantic meanings of the textual sequence.

\subsubsection{Gradient Leakage}
Gradient leakage commonly refers to recovering input texts given access to their corresponding model gradients.
For federated learning, the FedAvg algorithm \cite{konevcny2016federated} requires access to gradients to update model parameters.
Thus, gradient leakage problems~\cite{Zhu-2019-Deep,Zhao-2020-iDLG} are widely studied in computer vision and still rather unexplored for natural language processing, especially for language models due to the discrete optimization.
\citet{balunovic2022lamp} used an auxiliary language model to model prior probability and optimized the reconstruction on embeddings.
\citet{gupta2022recovering} extended LMs' gradient leakage to larger batch sizes.
\citet{fowl2023decepticons} studied gradient leakage on the first transformer layer to construct malicious
parameter vectors.
\citet{chu2023panning} considered targeted sensitive patterns extraction and decoded them from aggregated gradients.

Consequently, these studies on gradient leakage reveal that simple federated learning frameworks for LLMs are insufficient to support these frameworks' privacy claims.

\subsection{Others}
Beyond the above widely noticed privacy threats, here we also identify several related and under-explored potential privacy threats.

\subsubsection{Prompt Extraction Attacks}
Prompts are vital in LLMs' development to understand and follow human instructions.
Several powerful prompts enable LLMs to be smart assistants with external applications.
These prompts are of high value and are usually regarded as commercial secrets.
Prompt extraction attacks, also known as prompt stealing attacks, aim to recover the secret prompts via interactions with LLMs.
Prompt extraction attacks can be done via prompt injection attacks.
\citet{ignore_previous_prompt, Liu2023PromptIA} introduced the \textit{prompt injection} method, which enables the leakage of specially designed prompts in applications built upon LLMs.
To infer the precious prompts, prompt extraction attacks \cite{Zhang2023PromptsSN} were proposed to evaluate the effectiveness of the attack performance. \citet{zhang2024extracting} trained the prompt extraction model with the response-prompt datasets.

In terms of evaluation,
\citet{toyer2024tensor,schulhoff-etal-2023-ignore} gathered around 12.6k and 600k prompt extraction instances.
Furthermore, \citet{wang2024raccoon} proposed an evaluation benchmark to compare prompt extraction attacks' performance with 14 categories.

\subsubsection{Adversarial Attacks}
Adversarial attacks are commonly studied to exploit the models' instability to small perturbations to original inputs.
Several investigations~\cite{guo-etal-2021-gradient, Yang-2022-Natural,nguyen2023adversarial,wallace-etal-2021-concealed,Sadrizadeh2023TransFoolAA,gainski-balazy-2023-step,fang-etal-2023-modeling,wang2023adversarial,maus2023adversarial,lei-etal-2022-phrase} were conducted to learn LLMs' potential weaknesses.
Moreover, adversarial attacks against multi-modal LLMs were also recently examined~\cite{Carlini2023AreAN,Qi2023VisualAE}.

\textbf{LLM Red-teaming}.
LLM Red-teaming can be viewed as a variant or an extension of the adversarial attack.
The original concept of red-teaming emerged from adversarial simulations and war games conducted within military contexts.
For LLMs, Red-teaming is a rigorous security audit method designed to expose potential risks and failure modes through various challenging inputs~\cite{feffer2024redteaminggenerativeaisilver, ganguli2022redteaminglanguagemodels, yoo2024csrtevaluationanalysisllms}.
Given that manual red-teaming is costly and slow~\cite{touvron2023llama, bai2022training}, automatic methods are often more effective.
The most straightforward idea is to simulate the adversarial scenario and let two LLMs to attack and defense, respectively~\cite{ge-etal-2024-mart, jiang2024dartdeepadversarialautomated}.
To take a step forward for automatic red-teaming, \citet{deng-etal-2023-attack} instructed LLMs to mirror manual red-teaming prompts through in-context learning.
Similarly, \citet{yu2023gptfuzzer} mutated red-teaming templates rooted in human-written prompts. 
\citet{perez-etal-2022-red} explored the possibility of Reinforcement Learning (RL) \cite{Sutton1998ReinforcementLA} in automatic LLM red-teaming, leveraging the reward of a classifier to train a red-team LLM and then maximize the elicited harmfulness.
\citet{hong2024curiositydriven} further optimized the coverage of red-team prompts via a curiosity-driven RL method. 
An entropy bonus is incorporated to encourage the model's randomness, and a novelty reward is proposed for discovering unseen test cases.
In addition to these methods, \citet{bhardwaj2023redteaminglargelanguagemodels} utilized the Chain of Utterances (CoU) to reveal LLMs' internal thought, which significantly reduces model refusal rates.
\citet{chen2024agentpoisonredteamingllmagents} paid more attention to LLM-based agents, poisoning the Retrieval-Augmented Generation (RAG) knowledge base to elicit harmful responses.
Besides, several LLM red-teaming benchmarks and datasets have been published for further exploration~\cite{yoo2024csrtevaluationanalysisllms, radharapu-etal-2023-aart, liu2024mmsafetybenchbenchmarksafetyevaluation}.

\subsubsection{Side Channel Attacks}
\lhr{
Recently, \citet{debenedetti2023privacy} systematically formulated possible privacy side channels for systems developed from LLMs.
Four components of such systems, including training data filtering, input preprocessing, model output filtering and query filtering were identified as privacy side channels.
Given access to these four components, stronger membership inference attacks could be conducted via exploiting the design principles reversely.
}

\subsubsection{Decoding Algorithm Stealing}
The decoding algorithms with appropriate hyper-parameters contribute to the high-quality response generation.
Yet, great efforts are paid to select the suitable algorithm and its internal parameters.
To steal the algorithm with its parameters, stealing attack~\cite{naseh2023risks} was proposed with typical API access.
\citet{ippolito2023reverse} introduced algorithms that aim to differentiate between the two widely used decoding strategies, namely top-k and top-p sampling. Additionally, they proposed methods to estimate the corresponding hyper-parameters associated with each strategy.

\begin{figure}[t]
  \centering
  
  \includegraphics[width=0.92\linewidth]{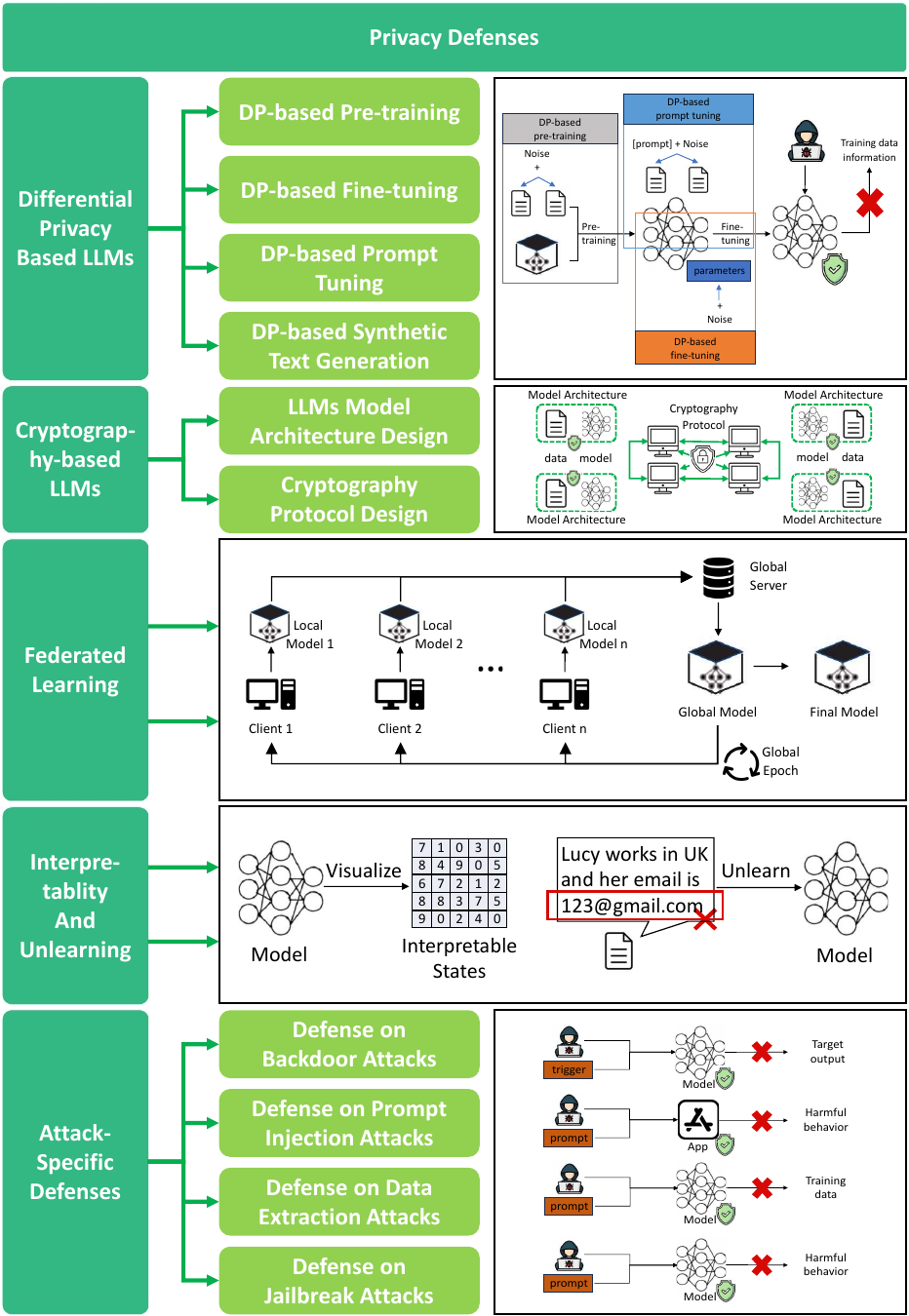}
  \vspace{-0.1in}
  \caption{An overview of existing privacy defenses on LLMs.}
  \label{fig:defense_overview}
  \vspace{-0.15in}
\end{figure}

\section{Privacy Defenses}\label{ch:Privacy Defenses}
To address the privacy attacks mentioned in Section~\ref{ch:Privacy Attacks}, in this section, we discuss existing privacy defense strategies to protect data privacy and enhance model robustness against privacy attacks.

\subsection{Differential Privacy Based LLMs}
As discussed in Section~\ref{sec:dp}, differential privacy enjoys several elegant mathematical properties, such as easy composition and post-processing.
Composition property allows an easy combination of multiple differential privacy mechanisms and helps to accumulate privacy budgets.
Post-processing guarantees data privacy after DP mechanism no matter how the privatized data is further processed.
Simple formulation with user-friendly properties makes differential privacy widely used across multiple fields for data protection and popular among Internet and technology companies.

For deep learning, noisy optimization algorithms such as DPSGD~\cite{Abadi-dpsgd-2016} enable model training with differential privacy guarantees.
DPSGD injects per-sample Gaussian noise with given noise scales into computed gradients during optimization steps and can be easily incorporated into various models.
Hence, most related works on privacy-preserving LLMs are developed with DPSGD as the critical backbone.
In this section, we divide existing DP-based LLMs into four clusters, including DP-based pre-training, DP-based fine-tuning, DP-based prompt tuning and DP-based synthetic text generation.

\subsubsection{DP-based Pre-training}
Since DP mechanisms have different implementations on LLMs, DP-based pre-training could further enhance LMs' robustness against perturbed random noises.
\citet{yu2023selective} proposed selective pre-training with differential privacy to improve DP fine-tuning performance on BERT.
\citet{Igamberdiev-2023-DP-BART} implemented a DP-BART for text rewriting under LDP with and without pre-training.

\subsubsection{DP-based Fine-tuning}
Most LLMs are pre-trained on publicly available data and fine-tuned on sensitive domains. 
It is natural to fine-tune LLMs on sensitive domains directly with DPSGD.
\citet{Feyisetan-20-Privacy} applied $d\-{\chi}$ privacy~\cite{Alvim-2018-Local,Chatzikokolakis-2013-Broadening}, a variant of local differential privacy (LDP)~\cite{kasiviswanathan-2011-localdp}, on the word embedding space to perform text perturbation on Bi-LSTM.
Similarly, \cite{qu2021natural} applied $d_\chi$ privacy to pre-train and fine-tune BERT.
While fine-tuning with noisy token embeddings leads to the utility degradation~\cite{qu2021natural}, \citet{du2023sanitizing} considered the perturbation of sentence embeddings for sequence-level metric-LDP~\cite{alvim2018local}.
\citet{shi-etal-2022-just} proposed selective DP to apply differential privacy only on sensitive textual parts and applied it on RoBERTa and GPT-2. 
\citet{yu2022differentially} applied DPSGD on fine-tuning BERT and GPT-2 via several fine-tuning algorithms.
\citet{mireshghallah2022differentially} considered knowledge distillation during private fine-tuning on BERT.
In contrast to DPSGD, which perturbs the gradients in back-propagation, \citet{du2023dp} proposed perturbing the embeddings in the forward process to guarantee privacy.
\citet{huang2023privacy} considered privacy in retrieval-based language models, which will answer user questions based on facts stored in domain-specific datastore. They consider a scenario in which the domain-specific datastore is private and may contain sensitive information that should not be revealed.

\subsubsection{DP-based Prompt Tuning}
For generative LLMs, due to their colossal model sizes, parameter-efficient tuning methods such as prompt tuning are widely adopted to tune models on various downstream tasks~\cite{li-liang-2021-prefix,
lester-2021-prompttuning,
DBLP:journals/corr/abs-2309-08303,
DBLP:conf/acl/ChanLCLSWS23,
DBLP:conf/emnlp/JiangZW22,schick-2021-exploiting}.
Thus, it is imperative to study these efficient tuning methods with DP optimizers for LLMs.
\citet{Ozdayi2023} considered soft-prompt based prefix tuning methods on LLMs for both prompt-based attack and prompt-based defense under the training data extraction scope.
\citet{li2023privacy} proposed differentially private prompt-tuning methods~\cite{lester-2021-prompttuning,li-liang-2021-prefix} and evaluated the privacy on the embedding-level information leakage via attribute inference and embedding inversion attacks.
\citet{duan2023flocks}  also proposed differentially private prompt-tuning methods with DPSGD and PATE.
Here, the PATE~\cite{nicolas_pate,yoon2018pategan} is the abbreviation of Private Aggregation of Teacher Ensembles implemented on the Generative Adversarial Nets (GAN) framework.

\subsubsection{DP-based Synthetic Text Generation}
Generative LLMs naturally can generate multiple responses via sampling-based decoding algorithms.
For the DP-tuned LLMs, sampling texts from LLMs satisfy the post-processing theorem and preserve the same privacy budget.
\citet{yue2022synthetic} applied DPSGD for synthetic text generation and evaluated the performance based on canary reconstruction.
These synthetic texts can be obtained via conditional generation on LLMs and can be safely released to replace the original private data for other downstream tasks.
Similarly, \citet{mattern-2022-differentially} used DP optimizer to fine-tune GPT-2 for conditional synthetic text generation and evaluated privacy on duplication.
\citet{Tian-2022-SeqPATE} applied another DP mechanism, PATE, during GPT-2's sentence completion.

\subsection{Cryptography-based LLMs}
Cryptography techniques such as Secure Multi-Party Computation (SMPC) and Homomorphic Encryption (HE) are widely applied in the inference phase of large language models to protect the privacy of model parameters and inference data. 
However, directly using cryptography techniques for privacy-preserving inference in LLMs is highly inefficient. 
This inefficiency arises because privacy-preserving computations based on cryptography techniques incur additional computational and communication overhead compared to plaintext computations. 
For instance, using SMPC to perform large-scale matrix multiplications and non-linear operations (e.g., Softmax, GeLU, and LayerNorm) in LLMs requires significant communication costs. 
To improve the efficiency of cryptography techniques in privacy-preserving LLMs inference and enable scalability to larger models, current research efforts are divided into two main directions: LLMs model architecture design and cryptography protocol design.

\subsubsection{LLMs Model Architecture Design}
The LLM Model Architecture Design (MAD) approach aims to improve inference efficiency by leveraging the robustness of LLMs and modifying their architecture. Specifically, MAD involves replacing non-linear operations that are incompatible with SMPC and HE, such as Softmax, GeLU, and LayerNorm, with other operators that are compatible with these cryptographic techniques.

As an early work on privacy-preserving LLMs inference, \cite{chen2022x} presented an innovative implementation of privacy-preserving inference for the BERT model, utilizing homomorphic encryption~\cite{paillier1999public}. 
THE-X utilizes approximation methods such as polynomials and linear neural networks to replace non-linear LLM operations with addition and multiplication operations that homomorphic encryption can compute.
However, THE-X has the following three limitations: 1) It does not have provable safety. This is because in THE-X, clients need to decrypt the intermediate calculation results and complete the ReLU calculation in plain text; 2) Performance degradation caused by changes in the model structure; 3) Due to changes in model structure, retraining is required to adapt to the new model structure. 

To address these challenges, several researchers have explored using Secure Multi-Party Computation (SMPC) technologies, such as secret-sharing, to develop privacy-preserving algorithms for LLM inference. Notably, MPCFormer \cite{li2022mpcformer} proposed an approach that replaced the non-linear operations in the LLMs model with polynomials while leveraging model distillation to maintain performance. They validated the effectiveness of their algorithm through experiments conducted on multiple datasets, evaluating it across three scales of BERT models. Building upon this work, \citet{zeng2022mpcvit} incorporated principles from \cite{li2022mpcformer}'s approach and integrated Neural Architecture Search (NAS) technology to further enhance model efficiency and performance. Additionally, \citet{liang2023merge} integrated techniques from previous works \cite{chen2022x,li2022mpcformer}, specifically focusing on Natural Language Generation (NLG) tasks. To enhance the efficiency of privacy-preserving inference, they customized techniques like Embedding Resending and Non-linear Layer Approximation Fusion to better align with the inference characteristics of NLG models. These adaptations have proven highly effective in optimizing the efficiency of privacy-preserving inference for NLG tasks.

\subsubsection{Cryptography Protocol Design}
The Cryptography Protocol Design (CPD) refers to designing more efficient cryptography protocols to improve the efficiency of privacy-preserving inference in LLMs while maintaining the original model structure. For example, customized SMPC protocols such as Softmax, GeLU, and LayerNorm can be explicitly designed for non-linear LLM operations. Since the model structure remains unchanged, the performance of privacy-preserving inference using CPD-based LLMs is not affected compared to the plaintext model.

As the first work, \citet{hao2022iron} improved the efficiency of LLMs privacy protection model inference by integrating SMPC and HE. Specifically, it used HE to accelerate linear operations in LLMs privacy-preserving inference, such as matrix multiplication. For the non-linear operations, it used SS and look-up-table to design efficient privacy-preserving index and division algorithms, respectively. Different from \cite{hao2022iron}, \citet{zheng2023primer} used a confusion circuit (GC) to optimize the non-linear operation in LLM. \citet{sigma} constructed a secure calculation protocol for each function of LLMs based on function secret sharing (FSS), which greatly improved the efficiency of LLMs' privacy-preserving inference.

In addition, some studies \cite{dong2023puma, hou2023ciphergpt, ding2023east, luo2024secformer, pang2024bolt} proposed using piecewise polynomials or Fourier series to approximate the non-linear operators in LLMs, thereby improving inference efficiency. For example, \citet{dong2023puma} used piecewise polynomials to accurately approximate exponential and GeLU operations in LLMs and, through a series of engineering optimizations, achieved privacy-preserving inference for large-scale LLMs, such as LLaMA-7B. \citet{luo2024secformer} proposed using Fourier series to approximate the error function in the GeLU function, further enhancing the efficiency of private computation for the GeLU function. \citet{hou2023ciphergpt} proposed a preprocessing packaging optimization method for unbalanced matrix multiplication based on subfield-VOLE for the GPT model, which greatly reduced the preprocessing overhead of matrix multiplication. For non-linear processing, it adopted the piecewise fitting technique and followed SIRNN \cite{rathee2021sirnn} to optimize the computational efficiency of approximate polynomials.

\subsection{Federated Learning}

Federated learning (FL) is a privacy-preserving distributed learning paradigm~\cite{yangfml}, and it can be leveraged to enable multiple parties to train or fine-tune their LLMs collaboratively without sharing private data owned by participating parties~\cite{fan2023fatellm,kang2023grounding}.

While FL can potentially protect data privacy by preventing adversaries from directly accessing private data, a variety of research works have demonstrated that FL algorithms without adopting any privacy protection have the risk of leaking data privacy under data inference attacks mounted by \textit{semi-honest} ~\cite{Zhu-2019-Deep,Zhao-2020-iDLG,yin2021see,geiping2020inverting,gupta2022recovering,balunovic2022lamp} or \textit{malicious} adversaries ~\cite{fowl2023decepticons,chu2023panning}. Semi-honest adversaries follow the federated learning protocol but may infer the private data of participating parties based on observed information, while malicious adversaries may update intermediate training results or model architecture maliciously during the federated learning procedure to extract the private information of participating parties. The literature has explored various approaches to protect data privacy for LLMs' pre-training, fine-tuning, and inference. 




          


Pre-trained LLMs can be used to initialize clients' local models for better performance and faster convergence than random initialization. ~\citet{hou2023FreD} proposed FedD that fine-tunes an LLM using a public dataset adapted to private data owned by FL clients and then dispatched this LLM to clients for initialization. FedD collected statistical information on clients' private data through differentially private federated learning and leveraged this statistical information to select samples close to the distribution of clients' private data from the public dataset. Similarly, \citet{wang2023can} initialized clients' models from an LLM distilled from a larger LLM using public data adapted to clients' private data through a privacy-preserving distribution matching algorithm based on DP-FTRL (Follow-The-Regularized-Leader). 

For fine-tuning clients' local LLMs in the FL setting, ~\citet{zhang2023fedpetuning} proposed FedPETuning that fine-tunes clients' local models leveraging Parameter-Efficient-FineTuning (PEFT) techniques and demonstrated that federated learning combined with LoRA~\cite{hu2022lora} achieved the best privacy-preserving results among all compared PEFT techniques. \citet{xu2023lvnlm} proposed to combine DP with Partial Embedding Updates (PEU) and LoRA to achieve better privacy-utility-resource trade-off than baselines.

In federated transfer learning~\cite{kang2023grounding}, clients transfer knowledge from the server's LLM to their local models. Specifically, clients use their proprietary data as demonstrations to prompt the LLM, generating responses that may include reasoning explanations, rationales, or instructions. These responses are then used to fine-tune the clients' local models. To protect the privacy of the client's local data sent to the server, ~\citet{fan2024pdss} proposed employing data randomization to obscure the prompts sent to the server. Alternatively, \citet{li2024fdkt} introduced a method involving a client-side privacy-preserving generator that creates synthetic data, which is then forwarded to the server for inference, thereby preserving the client's data privacy.

\citet{gupta2022recovering}, \citet{balunovic2022lamp}, \citet{fowl2023decepticons} and \citet{chu2023panning} investigated the privacy attacks and defenses in federated LLM settings. \citet{gupta2022recovering} and 
\citet{balunovic2022lamp} proposed FILM and LAMP to recover a client's input text from its submitted gradients, and they leveraged FWD (freezing word
embeddings) and DP-SGD, respectively, to defend against proposed text reconstruction attacks. These two works focus on the semi-honest setting, while \citet{fowl2023decepticons} and \citet{chu2023panning} proposed Decepticons and Panning that aimed to recover input text of clients maliciously. More specifically, both Decepticons and Panning involve a malicious server that sends malicious model updates to clients to capture private or sensitive information, and they suggest leveraging DP-SGD to mitigate privacy leakage by at a cost in utility.

\subsection{Interpretability and Unlearning}
Recent efforts have been made on the interpretability of LLMs, aiming to offer researchers valuable insights into the models' internal mechanisms~\cite{feng-etal-2018-pathologies, 10.1145/3580305.3599240, 9224153}.
This understanding is fundamental to identifying and mitigating potential risks associated with LLMs. 
\citet{zhou2024alignmentjailbreakworkexplain} investigated LLM safety mechanisms using weak classifiers on intermediate hidden states, revealing that ethical concepts are learned during pre-training and refined through alignment.
Their experiments across various model sizes demonstrate how jailbreaks disrupt the transformation of early unethical classifications into negative emotions, offering new insights into LLM safety and jailbreak techniques. 
\citet{zhang2024shieldlmempoweringllmsaligned} introduced an LLM-based safety detector trained on a large bilingual dataset, which offers customized detection rules and explanations for its decisions. 
The interpretability in LLM safety has not been adequately explored and needs further insights.

Another line of work investigates unlearning. The concept of unlearning \cite{liu2024rethinkingmachineunlearninglarge,chen2023unlearnwantforgetefficient,yao2024largelanguagemodelunlearning,pmlr-v235-pawelczyk24a,chakraborty2024crossmodalsafetyalignmenttextual} has also emerged as a crucial aspect of LLM safety and privacy.
It aims to selectively remove or modify specific knowledge or behaviors from trained models, addressing concerns about privacy, misinformation, bias, and potentially harmful content.
Early in 2015, \cite{7163042} proposed Machine Unlearning, enabling systems to efficiently forget specific data to protect users' privacy. 
\citet{liu-etal-2024-towards-safer} introduced Selective Knowledge negation Unlearning (SKU) to remove harmful knowledge from LLMs while preserving their utility on normal prompts.   
\citet{liu-etal-2024-towards-safer} first selectively isolated the harmful knowledge in the model parameters, then negating such knowledge for a safer model. 
\citet{zhang2024safeunlearningsurprisinglyeffective} proposed an unlearning-based defense mechanism against LLM jailbreak attacks. 
By directly removing harmful knowledge, \citet{zhang2024safeunlearningsurprisinglyeffective} reduced Attack Success Rate from 82.6\% to 7.7\% on Vicuna-7B~\cite{vicuna2023}, significantly outperforming safety-aligned fine-tuned models like Llama2-7B-Chat~\cite{touvron2023llama}.
Besides, \citet{lu2022quark} also combined unlearning with Reinforcement Learning to reduce LLMs' toxicity, conditioning on reward tokens for a more controllable generation.

\subsection{Specific Defense}
The aforementioned defense methods are generally applicable and serve as systematic defenses. In this section, we present a detailed illustration of the defense mechanisms employed against specific attacks, including the backdoor and data extraction attacks.
\subsubsection{Defenses on Backdoor Attacks}
For deep neural networks (DNNs), different heuristic defense strategies are implemented to address backdoor attacks.
\citet{liu2018fine} proposed Fine-Pruning to defend against backdoor attacks on DNNs and \citet{chen2018detecting} proposed the Activation Clustering (AC) method to detect poisonous training samples designed.
\citet{hong2020effectiveness} uncovered common gradient-level properties shared among different forms of poisoned data and observed that poisoned gradients exhibited higher magnitudes and distinct orientations compared to clean gradients. Consequently, they proposed \textit{gradient shaping} as a defense strategy, leveraging DPSGD~\cite{Abadi-dpsgd-2016}. 

For NLP models, a small group of word-level trigger detection algorithms are proposed.
\citet{qi2020onion} proposed a straightforward yet effective textual backdoor defense called ONION, which utilized outlier word detection. In this approach, each word was assigned a score based on its impact on the sentence's perplexity. Words with scores surpassing the threshold were identified as trigger words.
\citet{chen2021mitigating} proposed a defense method called Backdoor Keyword Identification (BKI). BKI utilized functions to score the impact of each word in the text by analyzing changes in the internal neurons of LSTM. From each training sample, several words with high scores were selected as keywords. The statistical information of these keywords from all samples was then computed to identify the keywords belonging to the backdoor trigger sentence, referred to as backdoor keywords. By removing the poisoning samples that contain backdoor keywords from the training dataset, they could achieve a clean model through retraining.

In terms of present-day LLMs, new intuitions about preventing poisoned data are proposed.
\citet{wallace-etal-2021-concealed} noted that poisoned examples with \textit{no overlap} of the triggers often included phrases that lacked fluency in English.
Consequently,  these poisoned samples can be easily identified through perplexity analysis.
\citet{cui2022a} made an observation that the poisoned samples tended to cluster together and become distinguishable from the normal clusters. Motivated by this phenomenon, they introduced the \textit{CUBE}. 
This method utilizes an advanced density clustering algorithm known as HDBSCAN~\cite{mcinnes2017accelerated} to effectively identify the clusters of poisoned and clean data.
\citet{Wan2023Poisoning, wallace-etal-2021-concealed} suggested an early stopping strategy as a defense mechanism against poisoning attacks.
\citet{markov2023holistic} developed a comprehensive model to detect a wide range of undesired content categories, such as sexual content, hateful content, violence, self-harm, harassment, and their respective subcategories by utilizing Wasserstein Distance Guided Domain Adversarial Training (WDAT). \citet{shen2018wasserstein} encouraged the model to learn domain invariant representations.
Other works are inspired by inconsistent correlations between poisoned samples and their corresponding labels.
\citet{yan-etal-2023-bite} introduced DEBITE, an approach that effectively removed words with strong label correlation from the training set via calculating the z-score~\cite{gardner2021competency, wu2022generating}. 
Due to the fact that the poisoned samples were incorrectly labeled, \citet{Wan2023Poisoning} employed a training approach where the samples with the highest loss are identified as the poisoned ones.  \citet{li2024backdoor} proposed simulating the trigger, embedding it into the instruction, and training the backdoored model to generate clean responses.

\subsubsection{Defense on Prompt Injection Attacks}
Several defense strategies have been proposed \cite{sandwich_defense_2023, hines2024defending, willison_2023, chen2024struq, wallace2024instruction, yi2023benchmarking, piet2023jatmo, suo2024signed} to mitigate risks of prompt injection attacks. \citet{sandwich_defense_2023, yi2023benchmarking} recommend adding reminders to help ensure LLMs follow the original instructions. \citet{hines2024defending, willison_2023} propose using special tokens to mark the boundaries of valid data. \citet{piet2023jatmo} suggest training models focus only on specific tasks, which limits their ability to follow harmful instructions. \citet{chen2024struq, wallace2024instruction} advocate for fine-tuning LLMs on datasets that teach them to prioritize approved instructions. Finally, \citet{suo2024signed} introduce a method where instructions are signed with special tokens, ensuring that LLMs only follow those that are properly signed.

\subsubsection{Defenses on Data Extraction Attacks}
~\citet{patil2023can} proposed an attack-and-defense framework to investigate directly deleting sensitive information from model weights.
Firstly, ~\citet{patil2023can} examined two attack scenarios:  1) retrieving data from concealed representations (white-box) and 2) generating model-based alternative phrasings of the initial input utilized for model editing (black-box).
Then ~\citet{patil2023can} proposed model weights editing methods combined with six defense strategies to defend against data extraction attacks.

Given that privacy falls under the sub-topic of safety, techniques for filtering toxic output \cite{dathathri2019plug, gehman2020realtoxicityprompts, schick2021self, krause2020gedi, liu2021dexperts, xu2021detoxifying} can also be utilized to mitigate privacy concerns.
Approaches aimed at directly reducing the probability of generating toxic words \cite{gehman2020realtoxicityprompts, schick2021self} can help lower the likelihood of encountering privacy issues.
Sentence-level filtering methods, such as selecting the most nontoxic candidate from the generated options \cite{xu2021detoxifying}, can also take into consideration the level of privacy.

\begin{figure}
	\centering
	\begin{minipage}[ht]{1.0\linewidth}
		\centering
		\includegraphics[width=0.95\linewidth]{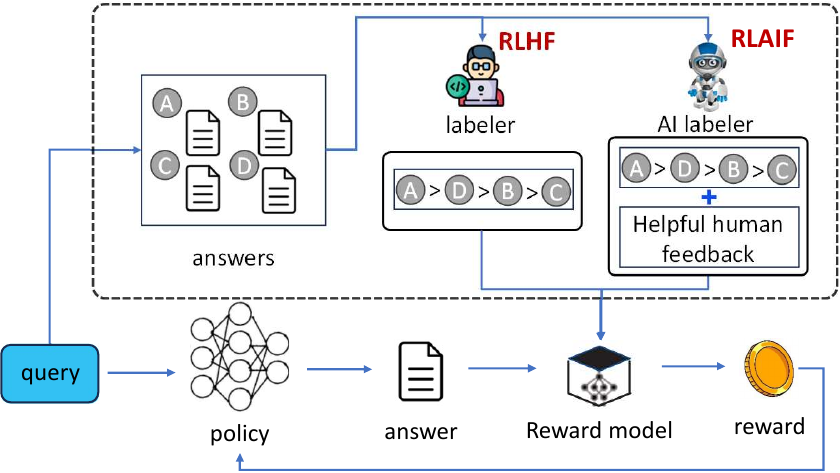}
  \vspace{-0.05in}
		\caption{Reinforcement Learning with Human Feedback (RLHF) and AI Feedback (RLAIF).}
		\label{fig:rlhf}
    \vspace{-0.2in}
	\end{minipage}
\end{figure}

Reinforcement learning from human feedback (RLHF) methods \cite{ziegler2019fine, stiennon2020learning, ouyang2022training, bai2022training,OpenAI2023GPT4TR, touvron2023llama} can be employed to assist models in generating more confidential responses (Figure~\ref{fig:rlhf}).
\citet{OpenAI2023GPT4TR} proposed rule-based reward models (RBRMs), which are a collection of zero-shot GPT-4 classifiers. The RBRMs were trained using human-written rubrics, aiming to reward the model to reject harmful requests during the RLHF process.
\citet{touvron2023llama} utilized context distillation \cite{askell2021general} to efficiently enhance safety capabilities of LLMs in RLHF.
Moreover, reinforcement learning from AI feedback (RLAIF) \cite{bai2022constitutional}, which ranks appropriate responses using AI with constitutional principles (Figure~\ref{fig:rlhf}), can help prevent privacy leakages by adhering to privacy principles.

\subsubsection{Defense on Jailbreak Attacks}

The rising prevalence of jailbreak attacks has exposed critical vulnerabilities in LLMs, therefore substantial research efforts focused on developing robust defenses to counter these increasingly threats~\cite{lu2024eraserjailbreakingdefenselarge,hasan2024pruningprotectionincreasingjailbreak,xu-etal-2024-safedecoding,wang-etal-2024-self,ji2024defendinglargelanguagemodels, robey2024smoothllmdefendinglargelanguage,wang2024mitigatingfinetuningbasedjailbreak}.
\citet{lu2024eraserjailbreakingdefenselarge} unlearned harmful knowledge in LLMs while retaining general capabilities to defend against jailbreak attacks.
\citet{hasan2024pruningprotectionincreasingjailbreak} introduced WANDA pruning \cite{sun2024a} to enhance LLMs' resistance to jailbreak attacks without fine-tuning, while maintaining performance on standard benchmarks.
In addition, \citet{hasan2024pruningprotectionincreasingjailbreak}  believed that the improvements can be understood through a regularization perspective.
As the opposite of GCG~\cite{zou2023universaltransferableadversarialattacks} that maximize the possibility of affirmative response for vicious prompts, \citet{xu-etal-2024-safedecoding} proposed to maximize the possibility of benign tokens with a safe-decode strategy. 
\citet{wang-etal-2024-self} combined safety training and safeguards to defense jailbreak attacks.
By training LLMs to self-review and tag their responses as harmful or harmless,\citet{wang-etal-2024-self} leverages the model's capabilities for harm detection while maintaining flexibility and performance. 
\citet{ji2024defendinglargelanguagemodels, robey2024smoothllmdefendinglargelanguage} transformed the given input prompts into different variants, and aggregated the responses for these inputs to detect adversarial inputs. 
\citet{wang2024mitigatingfinetuningbasedjailbreak} introduced a backdoor-enhanced safety alignment method to counter jailbreak attacks in Language-Model-as-a-Service (LMaaS) settings, effectively safeguarding LLMs with minimal safety examples by using a secret prompt as a backdoor trigger.

The shift towards Multimodal LLMs (MLLMs) introduces new vulnerabilities for potential jailbreaks, leading to the corresponding defense against multimodal jailbreak attacks.
\citet{wang2024adashieldsafeguardingmultimodallarge} proposed an adaptive prompting method to defend MLLMs against structure-based jailbreak attacks. 
A static defense prompt and an adaptive auto-refinement framework are utilized to improve MLLMs' robustness without fine-tuning or additional modules, while maintaining performance on benign tasks.
\citet{wang2024inferalignerinferencetimealignmentharmlessness} extracted safety-oriented vectors from aligned models to adjust the target model's internal representations, thus steering the model towards generating safe and appropriate responses when processing potentially harmful prompts.
\citet{zhang2024jailguarduniversaldetectionframework} proposed a universal detection framework for jailbreak and hijacking attacks on LLMs and MLLMs. 
It mutated untrusted inputs into disparate variants, and leveraged response discrepancies to distinguish attack samples from benign ones.
\citet{pi2024mllmprotectorensuringmllmssafety} employed a harm detector to identify harmful responses and a detoxifier to transform them into harmless ones.
\citet{gou2024eyesclosedsafetyon} proposed a training-free approach by transforming unsafe images into text, thereby activating the inherent safety mechanisms of aligned LLMs.
\citet{chen2024bathe} propose treating harmful instructions as backdoor triggers, prompting the MLLMs to generate rejection responses.
\citet{chakraborty2024crossmodalsafetyalignmenttextual} adopted unlearning to eliminate harmful knowledge in models, and effectively reduce the Attack Success Rate (ASR) for both text-based and vision-text-based attacks.
Furthermore, \citet{chakraborty2024crossmodalsafetyalignmenttextual} demonstrated that textual unlearning is more efficient than multimodal unlearning, offering comparable safety improvements with significantly lower computational costs.

\section{Future Directions on Privacy-preserving LLMs}\label{ch:Future}
After a thorough review of existing privacy attacks and defenses, in this section, we discuss the promising future directions for privacy-preserving LLMs.
First, we point to existing limitations on privacy attacks and defenses.
Then, we propose several promising fields that are currently less explored.
Finally, we draw our conclusions to summarize all explored aspects of this survey.

\subsection{Existing Limitations}
In this section, we conclude the limitations of existing works from the perspectives of both the adversary and defenders.

\subsubsection{Impracticability of Privacy Attacks}

The fundamental philosophy of privacy attacks is that with more powerful accessibility, the adversary is expected to recover more sensitive information or gain more control over victim LLMs.
For example, given only the black-box model access, the adversary may conduct training data extraction attacks to recover a few training data.
In addition, if the adversary is given extra information, such as hidden representations or gradients, it is expected to recover the exact sensitive data samples based on the given extra information, such as attribute inference, embedding inversion, and gradient leakage attacks.

However, assumptions of a powerful adversary do not imply a high impact due to practical considerations.
For instance, white-box attacks assume the attackers can inspect and manipulate LLMs' whole training process.
Usually, these attacks are expected to achieve better attack performance.
However, present-day attacks still prefer to examine black-box attacks since white-box accesses are not granted in real-life scenarios.
Though Section~\ref{ch:Privacy Attacks} lists various kinds of advanced black-box privacy attacks towards pre-trained/fine-tuned LLMs, the motivations of a few attacks are still dubious. 
The assumption of modifying or injecting patterns into original inputs is not sound for some backdoor attacks and prompt injection attacks.
For attribute inference, embedding inversion and gradient leakage attacks mentioned in Section~\ref{ch: extra-info}, they can only justify their motivations within limited use cases such as federated learning and vector databases.
Moreover, the adversary's auxiliary dataset $D^{\text{aux}}$ is commonly assumed to be distributed similarly to the victim model's training/tuning data.
However, a similar distribution assumption may not be applicable to general cases.

In summary, these criticisms call for future research on privacy attacks during real-use cases.

\subsubsection{Limitations of Differential Privacy Based LLMs}\label{ch:limit-dp}
Currently, DP-tuned LLMs become mainstream to protect data privacy.
Unfortunately, DP still suffers from the following limitations.

\textbf{Theoretical worst-case bounding}. Differential privacy based LLMs, by definition, assume a powerful adversary that can manipulate the whole training data.
The privacy parameters, $(\epsilon,\delta)$, provide the worst-case privacy leakage bounding.
However, in real scenarios, the adversary is not guaranteed full control over LLMs' training data.
Hence, there is still a considerable gap between practical attacks and worst-case probabilistic analysis of privacy leakage according to differential privacy.

\textbf{Degraded utility}. DP tuning is usually employed on relatively small-scale LMs for particularly simple downstream datasets.
Though a few works claimed that with careful hyper-parameter tuning, DP-based LMs could perform similarly to normal tuning without DP on some downstream classification tasks.
However, most works still exhibited significant utility deterioration when downstream tasks became complicated.
The degraded utility weakens the motivation of DP-based fine-tuning.

\subsection{Future Directions}
After reviewing the existing approaches and limitations, in this section, we point out several promising future research directions.

\subsubsection{Ongoing Studies about Prompt Injection Attacks}
Prompt injection attacks have gained significant attention recently due to the impressive performance and widespread availability of large language models.
These attacks aim to influence the LLMs' output and can have far-reaching consequences, such as generating biased or misleading information, spreading disinformation, and even compromising sensitive data. 
As for now, several prompt injection attacks have been proposed to exploit vulnerabilities in LLMs and their associated plug-in applications.
Still, domain-grounded privacy and safety issues of LLMs' applications are relatively unexplored.

Moreover, as awareness about these attacks continues to grow, current safety mechanisms fail to defend against these new attacks.
Thus, it becomes increasingly urgent to develop effective defenses that enhance the privacy and security of LLMs.

\subsubsection{Future Improvements on Cryptography}
The field of privacy-preserving inference for LLMs using cryptography techniques has developed rapidly, with a plethora of related research emerging. Researchers in machine learning and privacy protection have pursued two main technical routes: Model Architecture Design (MAD) and Cryptographic Protocol Design (CPD). Each of these routes has distinct advantages. Generally, MAD-based privacy-preserving inference algorithms for LLMs improve efficiency by modifying the model structure to circumvent expensive nonlinear operations. However, these methods may face limitations regarding privacy-preserving performance and model generalization. Conversely, CPD enhances the efficiency of privacy-preserving inference for LLMs by optimizing cryptography protocols while retaining the original model structure. Although CPD ensures model performance and generalization, the efficiency improvements are relatively modest. How to integrate the two technical approaches, MAD and CPD, to design a privacy-preserving LLM inference algorithm that achieves a better balance between efficiency and performance is worth considering. \citet{luo2024secformer} has made some interesting attempts. However, advancing the practical application of cryptographic-based privacy-preserving inference algorithms for LLMs remains an ongoing research endeavor.

\subsubsection{Privacy Alignment to Human Perception}
Currently, most works on privacy studies concentrate on simple situations with pre-defined privacy formulation.
For existing commercial products, personally identifiable information is extracted via named entity recognition (NER) tools, and  PII anonymization is conducted before feeding it to LLMs.
These naive formulations exploit existing tools to treat all extracted pre-defined named entities as sensitive information.
On the one hand, these studies' privacy formulations may not always be accurate and accepted by everyone.
For instance, users may input fake personal information for story writing, and their requests cannot be satisfied if the phony information is anonymized or cleaned.
Additionally, if all locations are considered as PII and anonymized, LLMs may disappoint users who want to search for nearby restaurants.
On the other hand, these studies only cover a narrow scope and fail to provide a comprehensive understanding of privacy.
For individuals, our privacy perception is affected by social norms, ethnicity, religious beliefs, and privacy laws~\cite{Benthall-CI-2017, Shvartzshnaider-2016-Learning, fan2024goldcoin, li2024privacy}.
Therefore, different groups of users are expected to exhibit different privacy preferences.
However, such human-centric privacy studies remain unexplored.

\subsubsection{Empirical Privacy Evaluation}
For privacy evaluation, the most direct approach is to give out DP parameters for DP-tuned LMs.
This simple evaluation approach is commonly adopted for DP-based LMs.
As discussed in Section~\ref{ch:limit-dp}, DP provides the worst-case evaluation results.
Such worst-case results may not be appropriate for quantifying privacy leakage during intended model usage.
Several works~\cite{li2023privacy, Ozdayi2023, Li2023PBenchAM} started to use empirical privacy attacks as privacy evaluation metrics.
For instance, \citet{Li2023PBenchAM} proposed PrivLM-Bench to systematically and empirically evaluate LMs of various settings, including existing training data extraction attacks, membership inference attacks, and embedding inversion attacks.
Experimental results from P-Bench suggest that there is a huge gap between the actual attack performance and the imagined powerful attacker capability from defenders' perspectives.
Hence, proper trade-off studies and balances between attacks and defenses are expected for future work.


\subsubsection{Towards Contextualized Privacy Judgment}
In addition to case-specific privacy studies, a general privacy violation detection framework is still missing.
Current works are constrained in simplified scenarios including PII cleaning and removal of a single data sample.
Laborious efforts are paid to extract and redact sensitive PII patterns based on information extraction algorithms for both academic researchers and industrial applications.
However, even if sensitive data cleaning can be perfectly done, personal information leakage can still occur in the given context.
For instance, during multi-turn conversations with LLMs-based chatbots, it is possible to infer personal attributes based on the whole context even if every utterance of conversations covers no private information~\cite{staab2023beyond}.
What's more, users may also make up fake PII that no one's private information is included.
To solve such complex problems, privacy judgment frameworks with reasoning and context understanding ability in a long context should be examined.

\section{Conclusion}\label{ch:Conclusion}


In conclusion, this survey offers a detailed and comprehensive analysis of the evolving landscape in the context of language models and large language models. 
Through the exhaustive review of the current literature for more than 200 papers, we have identified crucial aspects of LLMs' privacy vulnerabilities and investigated emerging defense strategies to mitigate these risks. 
Although several innovative defense strategies have been proposed, it's clear that they are not yet adequate to achieve privacy-preserving LLMs, leaving a few attacks unaddressed for further investigation. Moreover, some defenses are easily bypassed, and our review reveals potential limitations of the LLM privacy defense mechanism.
For future works, we call for more attention to more comprehensive privacy evaluation and judgment to align privacy with practical impacts. 
In the end, we point out the promising research directions that hold the potential to advance this field significantly.  
We hope that our survey will serve as a catalyst for future research and collaboration, ensuring that technological advancements are aligned with stringent data security and privacy standards and regulations.

\newpage


\bibliographystyle{ACM-Reference-Format}
\bibliography{ref}



\end{document}